%% file: submission.tex
\documentclass[runningheads]{llncs}
\usepackage{graphicx}
\usepackage{amsmath,amssymb} 
\usepackage{color}
\usepackage{epsfig}
\usepackage{graphicx}
\usepackage{amsmath}
\usepackage{amssymb}
\usepackage{xcolor}
\usepackage{booktabs}
\usepackage{multirow}
\usepackage{color}
\usepackage{wrapfig}
\usepackage{xspace}
\usepackage{caption}
\usepackage{xspace}

\newcommand*{\etal}{et al.\@\xspace}

\newcommand{\myparagraph}[1]{\vspace{0.1em}\noindent\textbf{#1}}

\begin{document}
\pagestyle{headings}
\mainmatter

\title{A Hybrid Model for Identity Obfuscation by Face Replacement}
\authorrunning{Sun et al. ECCV'18}

\author{Qianru Sun \quad Ayush Tewari\quad Weipeng Xu\\ 
\quad Mario Fritz \quad Christian Theobalt \quad Bernt Schiele}
\institute{Max Planck Institute for Informatics, Saarland Informatics Campus\\
\small  {\email{\{qsun, atewari, wxu, mfritz, theobalt, schiele\}@mpi-inf.mpg.de}}}

\maketitle

\begin{abstract}
As more and more personal photos are shared and tagged in social media, avoiding privacy risks such as unintended recognition, becomes increasingly challenging. We propose a new hybrid approach to obfuscate identities in photos by head replacement. Our approach combines state of the art parametric face synthesis with latest advances in Generative Adversarial Networks (GAN) for data-driven image synthesis. On the one hand, the parametric part of our method  gives us control over the facial parameters and allows for explicit manipulation of the identity. On the other hand, the data-driven aspects allow for adding fine details and overall realism as well as seamless blending into the scene context. In our experiments we show highly realistic output of our system that improves over the previous state of the art in obfuscation rate while preserving a higher similarity to the original image content. 
\end{abstract}

\input{3-framework}

\input{5-experiment}

\section*{Acknowledgments}
This research was supported in part by German Research Foundation (DFG CRC 1223) and the ERC Starting Grant CapReal (335545). We thank Dr. Florian Bernard for the helpful discussions.

\bibliographystyle{splncs}
\bibliography{egbib}
\input{9_supple}
\end{document}

%% file: 3-framework.tex
\section{Introduction}
Visual data is shared publicly at unprecedented scales through social media.
At the same time, however, advanced image retrieval and face recognition algorithms, enabled by deep neural networks and large-scale training datasets, allow to index and recognize privacy relevant information more reliably than ever.
To address this exploding privacy threat, methods for reliable identity obfuscation are crucial. Ideally, such a method should not only effectively hide the identity information but also preserve the realism of the visual data, i.e., make obfuscated people look realistic. 

Existing techniques for identity obfuscation have evolved from simply covering the face with often unpleasant occluders, such as black boxes or mosaics, to more advanced methods that produce natural images~\cite{sun2018cvpr,joon17iccv,Sharif16AdvML}.
These methods either perturb the imagery in an imperceptible way to confuse specific recognition algorithms~\cite{joon17iccv,Sharif16AdvML}, or substantially modify the appearance of the people in the images, thus making them unrecognizable even for generic recognition algorithms and humans~\cite{sun2018cvpr}.
Among the latter category, recent work~\cite{sun2018cvpr} leverages a generative adversarial network (GAN) to inpaint the head region conditioned on facial landmarks. It achieves state-of-the-art performance in terms of both recognition rate and image quality.
However, due to the lack of controllability of the image generation process, the results of such a purely data-driven method inevitably exhibit 
artifacts by inpainting faces of unfitting face pose, expression or implausible shape. 
In contrast, parametric face models~\cite{tewari2017mofa} give us complete control of facial attributes and have demonstrated compelling results for applications such as face reconstruction, expression transfer and visual dubbing~\cite{tewari2017mofa,thies2016face,GZCVVPT16}.
Importantly, using a parametric face model allows to control the identify of a person as well as to 
preserve attributes such as face pose and expression
by rendering and blending an altered face over the original image. 
However, this naive face replacement yields unsatisfactory results, since (1) fine level details cannot be synthesized by the model, (2) imperfect blending leads to unnatural output images and (3) only the face region is obfuscated while the larger head and hair regions, which also contain a lot of identity information, remain untouched.

In this paper, we propose a novel approach that combines a data-driven method and a parametric face model, and therefore leverages the best of two worlds.
To this end, we disentangle and solve our problem in two stages (see Fig.~\ref{Figure_framework_2stream_mainhead}):
In the first stage, we replace the face region in the image with a rendered face of a different identity.
To this end we replace the identity related component of the original person in the parameter vector of the face model while preserving attributes of original facial expression.
In the second stage, a GAN is trained to synthesize the complete head image given the rendered face and an obfuscated region around the head as conditional inputs.
In this stage, the missing region in the input is inpainted and fine grained details are added, resulting in a photo-realistic output image.
Our qualitative and quantitative evaluations show that our approach significantly outperforms the baseline methods on publicly available datasets with both lower recognition rate and higher image quality.

\section{Related work}
\myparagraph{Identity obfuscation.}
Blurring the face region or covering it with occluders, such as a mosaic or a black bar, are still the predominant techniques for visual identity obfuscation in photos and videos.
The performance of these methods in concealing identity against machine recognition systems has been studied in~\cite{joon16eccv} and~\cite{McPherson2016arxiv}.
They show that these simple techniques not only introduce unpleasant artifacts, but also become less effective due to the improvement of CNN-based recognition methods.
Hiding the identity information while preserving the photorealism of images is still an unsolved problem.
Only a few works have attempted to tackle this problem.

For \emph{target-specific} obfuscations, Sharif \etal~\cite{Sharif16AdvML} and Oh \etal~\cite{joon17iccv} used \emph{adversarial example} based methods which perturb the imagery in an imperceptible manner aiming to confuse specific machine recognition systems. 
Their obfuscation patterns are invisible to humans and the obfuscation performance is strong. However, obfuscation can only be guaranteed for target-specific recognizers.

To confuse \emph{target-generic} machine recognizers and even human recognizers, Brkic \etal~\cite{brkicCVPRW17} generated full body images that overlay with the target person masks. However, synthesized persons with uniform poses do not match scene context which leads to blending artifacts in final images.
The recent work of~\cite{sun2018cvpr} inpaints fake head images conditioned on the context and blends generated heads with diverse poses into varied background and body poses in social media photos.
While achieving state-of-the-art performance in terms of both recognition rate and image quality, the results of such a purely data-driven method inevitably exhibits artifacts like the change of attributes such as face poses and expressions.

\myparagraph{Parametric face models.}
Blanz and Vetter~\cite{Blanz1999} learn an affine parametric 3D Morphable Model (3DMM) of face geometry and texture from 200 high-quality scans. 
Higher-quality models have been constructed using more scans~\cite{booth20163d}, or by using information from in-the-wild images~\cite{Booth_2017_CVPR,tewari2017self}.
Such parametric models can act as strong regularizers for 3D face reconstruction problems, and have been widely used in optimization based~\cite{thies2016face,Roth:2016,Booth_2017_CVPR,Romdhani:2005vt,GVWT13,shi2014automatic} and learning-based~\cite{RichaSK2016,sela2017unrestricted,TranCVPR17,Richardson_2017_CVPR,Dou_2017_CVPR,kim17InverseFaceNet} settings.
Recently, a model-based face autoencoder (MoFA) has been introduced~\cite{tewari2017mofa} which combines a trainable CNN encoder with an expert-designed differentiable rendering layer as decoder, which allows for end-to-end training on real images.
We use such an architecture and extend it to reconstruct faces from images where the face region is blacked out or blurred for obfuscation.
We also utilize the semantics of the parameters of the 3DMM by replacing the identity-specific parameters to synthesize overlaid faces with different identities. 
While the reconstructions obtained using parametric face models are impressive, they are limited to the low-dimensional subspace of the models.
Many high-frequency details are not captured and the face region does not blend well with the surroundings in overlaid 3DMM renderings.
Some reconstruction methods go beyond the low-dimensional parametric models~\cite{cao2015real,GZCVVPT16,GVWT13,shi2014automatic,Richardson_2017_CVPR,sela2017unrestricted},
but most lack parametric control of the captured high-frequency details. 

\myparagraph{Image inpainting and refinement.}
We propose a GAN based method in the second stage to refine the rendered 3DMM face pixels for higher realism as well as to inpaint the obfuscated head pixels around the rendered face. 
In \cite{ShrivastavaPTSW17,mueller2017ganerated}, rendered images are modified to be more realistic by means of adversarial training. The generated data works well for specific tasks such as gaze estimation and hand pose estimation, with good results on real images. 
Raymond \etal~\cite{YehCLHD16} and Pathak \etal~\cite{contextEncoder} have used GANs to synthesize missing content conditioned on image context. Both of these approaches assume strong appearance similarity or connection between the missing parts and their contexts. 
Sun \etal\cite{sun2018cvpr} inpainted head pixels conditioned on facial landmarks. 
Our method, conditioned on parametric face model renderings, gives us control to change the identity of the generated face while also synthesizing more photo-realistic results.

\section{Face replacement framework}
\label{sec:framework}

We propose a novel face replacement approach for identity obfuscation that combines a data-driven method with a parametric face model.

\begin{figure}[htp]
  \centering
  \vspace{-0.3cm}
  \includegraphics[width=0.9\linewidth]{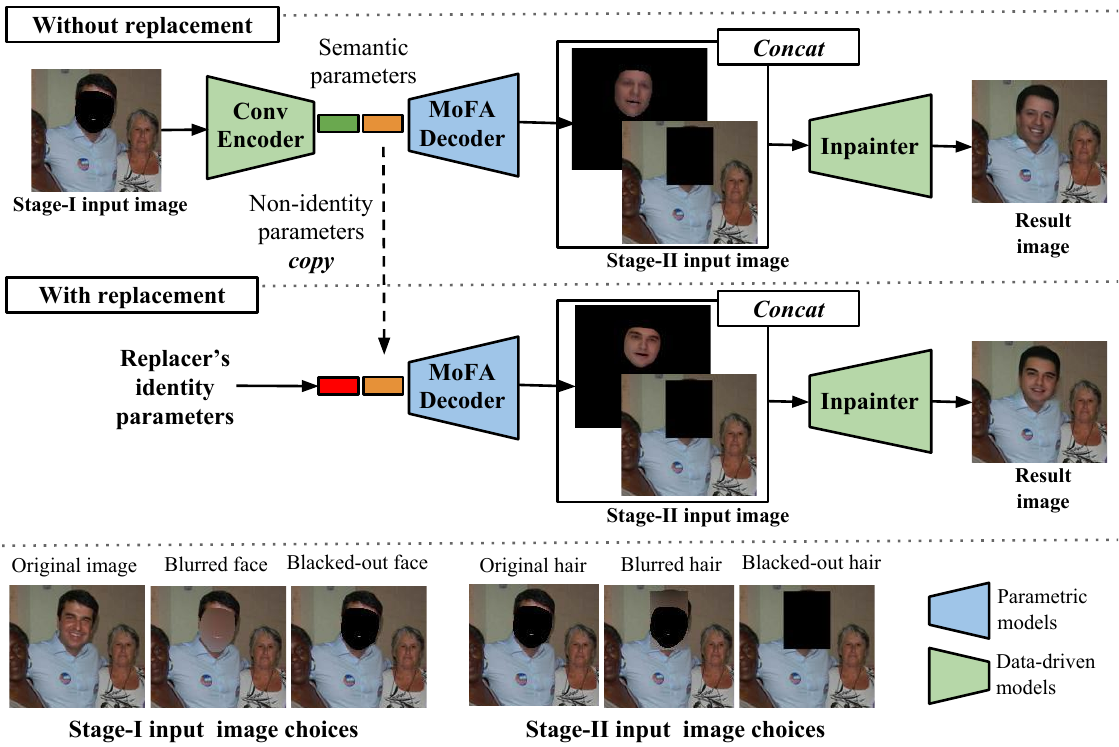}\\
  \caption{Our obfuscation method based on data-driven deep models and parametric face models.
  The bottom row shows the input image choices for stage-I and stage-II. Different input combination results in different levels of obfuscation.
  }
  \vspace{-0.6cm}
\label{Figure_framework_2stream_mainhead}
\end{figure}

Our approach consists of two stages (see Fig.~\ref{Figure_framework_2stream_mainhead}).
Experimenting on different modalities of input results in different levels of obfuscation~\footnote{Stage-I input image choices: \emph{Original image}, \emph{Blurred face} and \emph{Blacked-out face}.
Stage-II input image choices: \emph{Original hair},  \emph{Blurred hair} and \emph{Blacked-out hair}.}.
In the first stage, we can not only render a reconstructed face on the basis of a parametric face model (3DMM), but can also replace the face region in the image with the rendered face of a different identity.
In the second stage, a GAN is trained to synthesize the complete head image given the rendered face and a further obfuscated image around the face as conditional inputs. 
The obfuscation here protects the identity information contained in the ears, hair, etc.
In this stage, the obfuscated region is inpainted with realistic content and fine grain details missing in the rendered 3DMM are added, resulting in a photo-realistic output image.

\subsection{Stage-I: Face replacement}
Stage-I of our approach reconstructs 3D faces from the input images using a parametric face model. We train a convolutional encoder to regress the model's semantic parameters from the input.
This allows us to render a synthetic face reconstructed from a person and also gives us the control to modify its rendered identity based on the parameter vector.

\myparagraph{Semantic parameters.}
We denote the set of all semantic parameters as $p$ = $(\alpha, \beta, \delta, \phi, \gamma)$, $|p|=257$.
These parameters describe the full appearance of the face.
We use an affine parametric 3D face model to represent our reconstructions.
$\alpha$ and $\beta$ represent the shape and reflectance of the face, and correspond to the identity of the person. 
These parameters are the coefficients of the PCA vectors constructed from 200 high-quality face scans \cite{Blanz1999}.
$\delta$ are the coefficients of the expression basis vectors computed using PCA on selected blend shapes of \cite{Alexander2009} and \cite{Cao:2014}.
We use $80$ $\alpha$, $80$ $\beta$ and $64$ $\delta$ parameters. 
Together, they define the per-vertex position and reflectance of the face mesh represented in the topology used by \cite{tewari2017self}.
In addition, we also estimate the rigid pose ($\phi$) of the face and the scene illumination ($\gamma$). 
Rigid pose is parametrized with $6$ parameters corresponding to a $3D$ translation vector and Euler angles for the rotation. 
Scene illumination is parameterized using $27$ parameters corresponding to the first $3$ bands of the spherical harmonic basis functions \cite{Ramamoorthi2001}.

Our stage-I  architecture is based on the Model-based Face Autoencoder (MoFA)~\cite{tewari2017mofa} and consists of a convolutional encoder and a parametric face decoder.
The encoder regresses the semantic parameters $p$ given an input image\footnote{We use AlexNet\cite{krizhevsky2012imagenet} as the encoder.}. 

\myparagraph{Parametric face decoder.}
As shown in Fig.~\ref{Figure_framework_2stream_mainhead}, the parametric face decoder takes the output of the convolutional encoder, $p$, as input and generates the reconstructed face model and its rendered image.
The reconstructed face can be represented as $v_i(p) \in \mathbb{R}^{3}$ and $c_i(p) \in \mathbb{R}^{3}$, $\forall i \in [1, N]$, where $v_i(p)$ and $c_i(p)$ denote the position in camera space and the shaded color of the vertex $i$, and $N$ is the total number of vertices.
For each vertex $i$, the decoder also computes $u_i(p) \in \mathbb{R}^{2}$ which denotes the projected pixel location of $v_i(p)$ using a full perspective camera model. 

\myparagraph{Loss function.}
Our auto-encoder in stage-I 
is trained using a loss function that compares the input image to the output of the decoder as 
\begin{equation}
\label{eq:mofa_loss}
E_\textit{loss}(p) = E_\textit{land}(p) + w_\textit{photo} E_\textit{photo}(p) + w_\textit{reg} E_\textit{reg}(p).
\end{equation}
Here, $E_\textit{land}(p)$ is a landmark alignment term which measures the distance between 66 fiducial landmarks~\cite{tewari2017self} in the input image with the corresponding landmarks on the output of the parametric decoder, 
\begin{equation}
E_\textit{land}(p) = \sum_{i=1}^{66} || l_i - u_{x}(p) ||_2^2.
\end{equation}

$l_i$ is the  $i$th landmark's image position and $x$ is the index of the corresponding landmark vertex on the face mesh. 
Image landmarks are computed using the dlib toolkit~\cite{king2009dlib}.
$E_\textit{photo}(p)$ is a photometric alignment term which measures the per-vertex appearance difference between the reconstruction and the input image,
\begin{equation}
E_\textit{photo}(p) = \sum_{i \in \mathit{V}} || I(u_i(p)) - c_i(p) ||_2.
\end{equation}
$\mathit{V}$ is the set of visible vertices and $I$ is the image for the current training iteration. 
$E_{\textit{reg}}(p)$ is a Tikhonov style statistical regularizer which prevents degenerate reconstructions by penalizing parameters far away from their mean,
\begin{equation}
E_\textit{reg}(p) = \sum_{i=1}^{80} \frac{\alpha_i}{(\sigma_s)_i} + w_e \sum_{i=1}^{64} \frac{\delta_i}{(\sigma_e)_i} + w_r \sum_{i=1}^{80} \frac{\beta_i}{(\sigma_r)_i}.
\end{equation}
$\sigma_s$, $\sigma_e$, $\sigma_r$ are the standard deviations of the shape, expression and reflectance vectors respectively.
Please refer to \cite{tewari2017mofa,tewari2017self} for more details on the face model and the loss function.  
Since the loss function $E_{\textit{loss}}(p)$ is differentiable, we can backpropagate the gradients to the convolutional encoder, enabling self-supervised learning of the network.

\myparagraph{Replacement of identity parameters.}
\label{sec:replace_id}
The controllable semantic parameters of the face model have the advantage that we can modify them after face reconstruction. 
Note that the shape and reflectance parameters $\alpha$ and $\beta$ of the face model depend on the identity of the person~\cite{Blanz1999,TranCVPR17}. 
We propose to modify these parameters (referred to as identity parameters from now on) and render synthetic overlaid faces with different identities, while keeping all other dimensions fixed.
While all face model dimensions could be modified we want to avoid unfitting facial attributes. For example, changing all dimensions of the reflectance parameters can lead to misaligned skin color between the rendered face and the body.
To alleviate this problem, we keep the first, third and fourth dimensions of $\beta$, which control the global skin tone of the face, fixed.

\begin{figure}[htp]
\vspace{-0.4cm}
  \centering
  \includegraphics[width=0.95\linewidth]{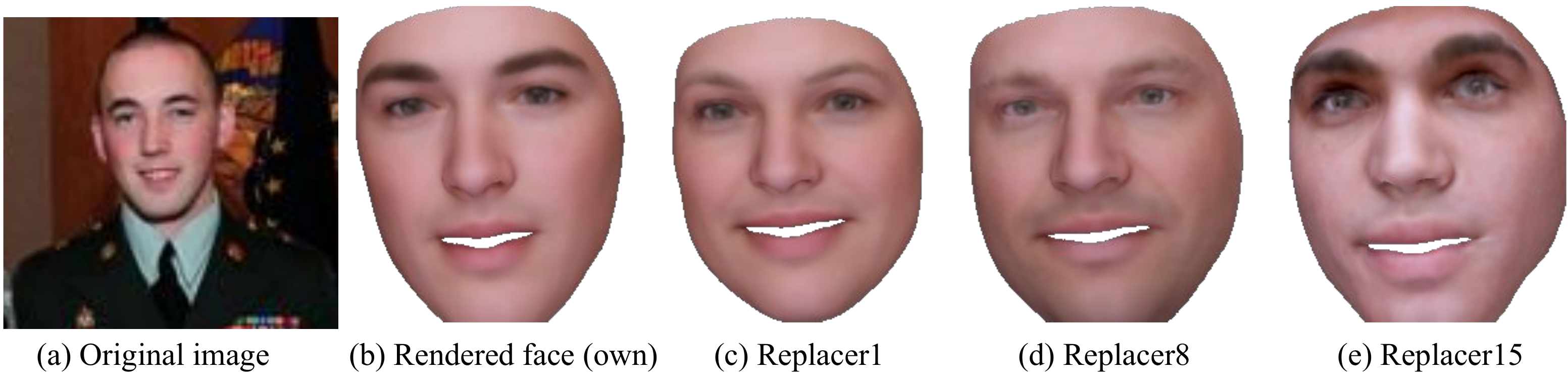}\\
  \caption{
  Replacement of identity parameters in Stage-I allows us to generate faces with different identities.}
   \vspace{-0.4cm}
\label{Figure_replacers}
\end{figure}

After obtaining the semantic parameters on all our training set (over 2k different identities), we first cluster the identity parameters into $15$ different identity clusters with the respective cluster means as representatives. 
We then replace the identity parameters of the current test image with the parameters of the cluster that is either closest (\texttt{Replacer1}), at middle distance (\texttt{Replacer8}) or furthest away (\texttt{Replacer15}) to evaluate different levels of obfuscation (Fig.~\ref{Figure_replacers}). Note that each test image has its own \texttt{Replacers}.

\begin{figure}[t]
	\centering
	\includegraphics[width=\linewidth]{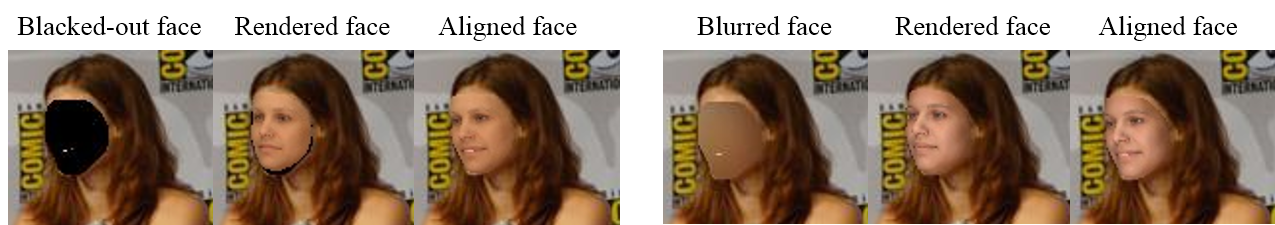}
	\caption
    {
	Stage-I output: If the face in the input image is blacked out or blurred, our network can still predict reasonable parametric reconstructions which align to the contour of the face region.
	The appearance is also well estimated from the context information.
    The results are further aligned using an optimization-based strategy. 
    }
     \vspace{-0.4cm}
	\label{fig:landmark_refinement}
	\end{figure}

\myparagraph{Input image obfuscation.} 
In addition to replacing the identity parameters, we also optionally allow additional obfuscation by blurring or blacking out the face region in the input image for Stage-I (the face region is determined by reconstructing the face from the original image).
These obfuscation strategies force the Stage-I network to predict the semantic parameters only using the context information (Fig.~\ref{fig:landmark_refinement}), thus reducing the extent of facial identity information captured in the reconstructions. 
We train networks for these strategies using the full body images with the obfuscated face region as input while using the original unmodified images in the loss function $E_{\textit{loss}}(p)$\footnote{If the input image is not obfuscated in Stage-I, we directly use the pre-trained coarse model of~\cite{tewari2017self} to get the parameters and the rendered face.}.
This approach gives us results which preserve the boundary of the face region and the skin color of the person even for such obfuscated input images (Fig. \ref{fig:landmark_refinement}).
The rigid pose and appearance of the face is also nicely estimated. 

In addition to reducing the identity information in the rendered face, the Stage-I network 
also removes the expression information when faces in the input images are blurred or blacked out. 
To  better align our reconstructions with the input images without adding any identity-specific information, we further refine only the rigid pose and expression estimates of the reconstructions.
We minimize part of the energy term in \eqref{eq:mofa_loss} after initializing all parameters with the predictions of our network.

\begin{equation}
p^{*} = \underset{p}{\mathrm{argmin}} \, E_\textit{refine}(p)
\end{equation}
\begin{equation}
E_\textit{refine}(p) = E_\textit{land}(p) + w_\textit{reg} E_\textit{reg}(p) 
\end{equation}
Note that only $\phi$ and $\delta$ are optimized during refinement. 
We use $10$ non-linear iterations of a Gauss-Newton optimizer to minimize this energy.
As can be seen in Fig \ref{fig:landmark_refinement}, this optimization strategy significantly improves the alignment between the reconstructions and the input images.
Note that input image obfuscation can be combined with identity replacement to further change the identity of the rendered face.

The output of stage-I is the shaded rendering of the face reconstruction.
The synthetic face lacks high-frequency details and does not blend perfectly with the image as the expressiveness of the parametric model is limited. 
Stage-II enhances this result and provides further obfuscation by removing/reducing the context information from the full head region.

\input{3-refinement}

%% file: 3-refinement.tex
\subsection{Stage-II: Inpainting}
\label{sec:inpainting}

Stage-II is conditioned on the rendered face image from Stage-I and an obfuscated region around the head to inpaint a realistic image. 
There are two objectives for this inpainter: (1) inpainting the blurred/blacked-out hair pixels in the head region; (2) modifying the rendered face pixels to add fine details and realism to match the surrounding image context.
The architecture is composed of a convolutional generator $G$ and discriminator $D$, and is optimized by L1 loss and adversarial loss.

\myparagraph{Input.} 
For the generator $G$, RGB channels of both the obfuscated image $I$ and the rendered face from Stage-I $F$ are concatenated as input. 
For the discriminator $D$, we take the inpainted image as \textit{fake} and the original image as \textit{real}. Then, we feed the (\textit{fake}, \textit{real}) pairs into the discriminator. 
We use the whole body image instead of just the head region in order to generate  natural transitions between the head and the surrounding regions including body and background, especially for the case of obfuscated input.

\myparagraph{Head Generator ($G$) and Discriminator ($D$).} The head generator $G$ is a ``U-Net''-based architecture~\cite{U-net}, i.e. Convolutional Auto-encoder with skip connections between encoder and decoder\footnote{Network architectures and hyper-parameters are given in supplementary materials.}, following \cite{sun2018cvpr}\cite{PG2}\cite{Ma2018}. It generates a natural head image given both the surrounding context and the rendered face. 
The architecture of the discriminator $D$ is the same as in DCGAN~\cite{Radford2015-DCGAN}.

\myparagraph{Loss function.} We use L1 reconstruction loss plus adversarial loss, named ${\cal{L}}^{G}$, to optimize the generator and the adversarial loss, named ${\cal{L}}^{D}$, to optimize the discriminator.  
For the generator, we add the head-masked L1 loss such that the optimizer focuses more on the appearance of the targeted head region,
\begin{equation}
	{\cal{L}}^{G} = {\cal{L}}_{bce}({{D}(G(I, F)),1}) + \lambda \| (G({I}, F) - I_O) \odot {M_h} \|_1 ,
    \label{eq:head_mask_L1_loss}
\end{equation}
where $M_h$ is the head mask (from the annotated bounding box), $I_O$ denotes the original image and ${\cal{L}}_{bce}$ is the binary cross-entropy loss. 
$\lambda$ controls the importance of the L1 term\footnote{When $\lambda$ is too small, the adversarial loss dominates the training and it is more likely to generate artifacts; when $\lambda$ is too big, the generator mainly uses the L1 loss and generates blurry results.}.
Then, for the discriminator, we have the following losses:
\begin{eqnarray}	
    {\cal{L}}^{D}={\cal{L}}^{D}_{adv} &=& {\cal{L}}_{bce}(D{(I_O),1}) + {\cal{L}}_{bce}({D(G(I, F)),0}), 
	\label{eq:adversarial_D_loss}
\end{eqnarray}

Note that de-identification losses derived from identity verification models \cite{MirjaliliRNR18} are not applicable to our approach. Since such verification models are usually trained by real images,
they are not able to distinguish the trivial difference between real and inpainted images. We observed that the verification loss becomes zero when the inpainted image is close enough to the ground truth image.

\begin{figure}[htp]
  \centering
  \includegraphics[width=0.8\linewidth]{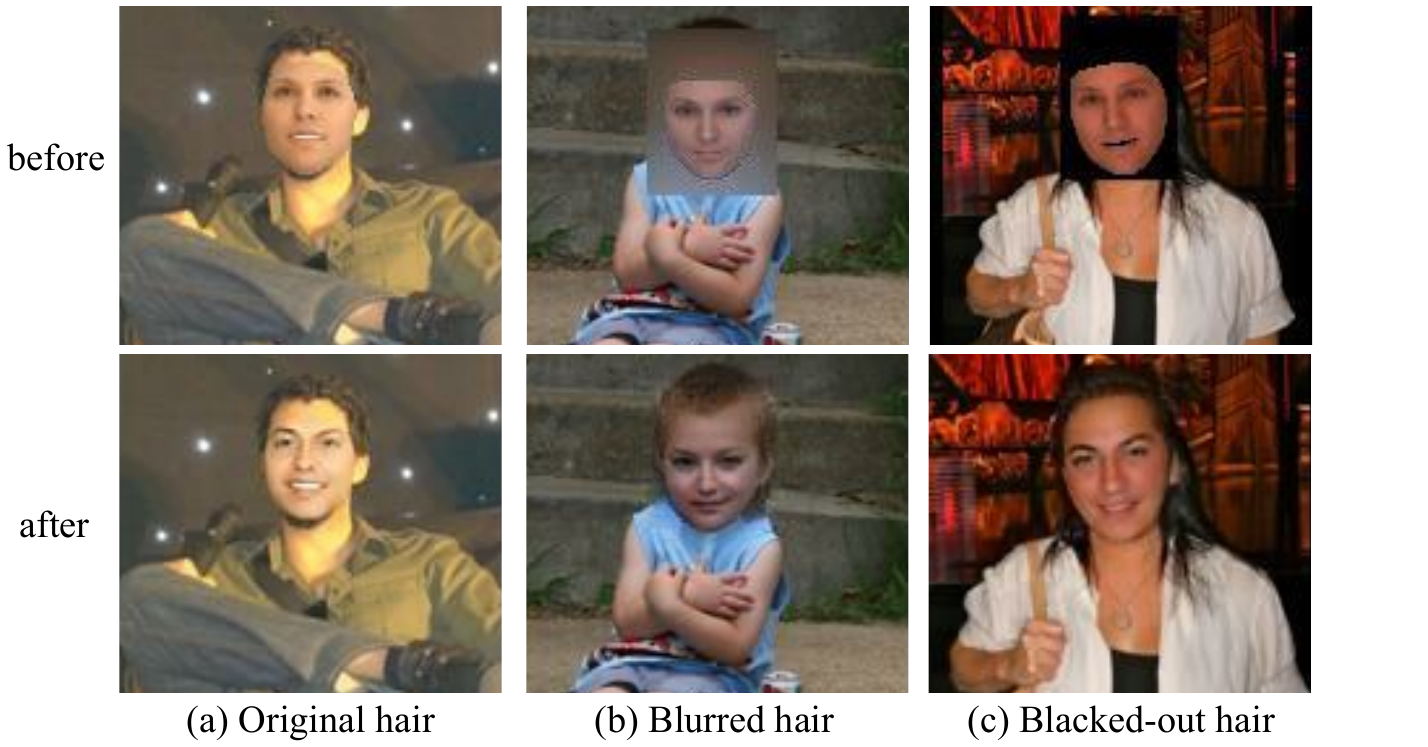}\\
  \caption{Visualization results before and after inpainting. On the top row, rendered faces are overlayed onto the color images for better comparison of details.}
   \vspace{-0.4cm}
\label{Figure_stageII_effect}
\end{figure}

Fig.~\ref{Figure_stageII_effect} shows the effect of our inpainter. In (a) when the original hair image is given, the inpainter refines the rendered face pixels to match surroundings, e.g., the face skin becomes more realistic in the bottom image. In (b)(c), the inpainter not only refines the face pixels but also generates the blurred/missing head pixels based on the context.

\vspace{-0.2cm}
\section{Recognizers}
\label{sec:two_recognizers}

Identity obfuscation in this paper is target-generic: it is designed to work against any recognizer, be it machine or human. In this paper, we use both recognizers to test our approach.

\subsection{Machine recognizers}
\vspace{-0.1cm}

We use an automatic recognizer \texttt{naeil} \cite{Oh2015Iccv}, the state-of-the-art for person recognition in social media images \cite{sun2018cvpr}\cite{JoonPAMIsub1710}.  
In contrast to typical person recognizers, \texttt{naeil} also uses body and scene context cues for recognition. It has thus proven to be relatively immune to common obfuscation techniques like blacking-out or blurring the head region~\cite{joon16eccv}.

We first train feature extractors over head and body regions, and then train SVM identity classifiers on those features. We can concatenate features from multiple regions (e.g. head+body) to make use of multiple cues. In our work, we use GoogleNet features from \texttt{head} and \texttt{head+body} for evaluation. We have also verified that the obfuscation results show similar trends against AlexNet-based analogues (see supplementary materials).

\vspace{-0.1cm}
\subsection{Human recognizers}
\vspace{-0.1cm}

We also conduct human recognition experiments to evaluate the obfuscation effectiveness in a perceptual way.
Given an original head image and the head images inpainted by variants of our method and results of other methods, we ask users to recognize the original person from the inpainted ones, and to also choose the farthest one in terms of identity. Users are guided to focus on identity recognition rather than the image quality. For each method, we calculate the percentage of times its results were chosen as the farthest identity (higher number implies better obfuscation performance).

%% file: 5-experiment.tex
\section{Experiments}

An obfuscation method should not only effectively hide the identity information but also produce photo-realistic results.
Therefore, we evaluate our results on the basis of recognition rate and visual realism.
We also study the impact of different levels of obfuscation yielded from different input modalities at two stages.

\subsection{Dataset}

Our obfuscation method needs to be evaluated on realistic social media photos. PIPA dataset~\cite{PIPA} is the largest social media dataset (37,107 Flickr images with 2,356 annotated individuals), which shows people in diverse events, activities and social relations~\cite{SunCVPR17}. In total, 63,188 person instances are annotated with head bounding boxes from which we create head masks. 
We split the PIPA dataset into a training set and a test set without overlapping identities, following \cite{sun2018cvpr}. In the training set, there are 2,099 identities, 46,576 instances and in the test set 257 identities, 5,175 instances. We further prune images with strong profile or back of the head views from both sets following \cite{sun2018cvpr}, resulting in 23,884 training and 1,084 test images.
As our pipeline takes a fixed-size input ($256\times 256\times 3$), we normalize the image size of the dataset.
To this end, we crop and zero-pad the images so that the face appears in the top middle block of a $3\times 4$ grid in the entire image. Details of our crop method are given in supplementary materials.

\subsection{Input modalities}

Our method allows 18 different combinations of input modalities, which is a combination of 3 types of face modalities, 3 types of hair modalities
and the choice of modifying the face identity parameters (default replacer is  \texttt{Replacer15}).
Note that, only 17 of them are valid for obfuscation, since the combination of original face and hair aims to reconstruct the original image. 
Due to space limitations, we compare a representative subset, as shown in Table~\ref{Table_quantitative}. 
The complete results can be found in the supplementary material.

In order to blur the face and hair regions in the input images, we use the same Gaussian kernel as in~\cite{sun2018cvpr,joon16eccv}. 
Note that in contrast to those methods, our reconstructed face model provides the segmentation of the face region allowing us to precisely blur the face or hair region.

\subsection{Results}

In this section, we evaluate the proposed hybrid approach with different input modalities in terms of the realism of 
images and the obfuscation performance.

\myparagraph{Image realism.}
We evaluate the quality of the inpainted images compared to the ground truth (original) images using Structure Similarity Score (SSIM) ~\cite{ImageQuality}. During training, the body parts are not obfuscated, so we report the mask-SSIM~\cite{sun2018cvpr,PG2} for the head region only (SSIM scores are in supplementary materials). This score measures how close the inpainted head is to the original head.

The SSIM metric is not applicable when using a \texttt{Replacer},
as ground truth images are not available.
Therefore, we conduct a human perceptual study (HPS) on Amazon Mechanical Turk (AMT) following~\cite{sun2018cvpr,PG2}. For each method, we show 55 real and 55 inpainted images in a random order to 20 users, who are asked to answer whether the image looks real or fake within 1s.

\myparagraph{Obfuscation performance.}
Obfuscation evaluation is to measure how well our methods can fool automatic person recognizers as well as humans. We have defined machine recognizers and human recognizers in Section~\ref{sec:two_recognizers}. 

For machine recognizers, we report the average recognition rates for $1,084$ test images in Table~\ref{Table_quantitative}. For human recognition, we randomly choose 45 instances then ask recognizers to verify the identity, given the original image as reference, from the obfuscated images of six representative methods: two methods in~\cite{sun2018cvpr} and four methods indexed by \textbf{v9}-\textbf{v12}, see the last column of Table~\ref{Table_quantitative}.

\input{5-Quantitative_results}

\section{Conclusion}
We have introduced a new hybrid approach to obfuscate identities in photos by head replacement.
Thanks to the combination of a parametric face model reconstruction and rendering, and the GAN-based data-driven image synthesis, our method gives us complete control over the facial parameters for explicit manipulation of the identity, and allows for photo-realistic image synthesis.
The images synthesized by our method confuse not only the machine recognition systems but also humans.
Our experimental results have demonstrated  output of our system that improves over the previous state of the art in obfuscation rate while generating obfuscated images of much higher visual realism.

%% file: 5-Quantitative_results.tex
\input{5-main_table}

\begin{figure}[htp]
  \centering
  \includegraphics[width=0.85\linewidth]{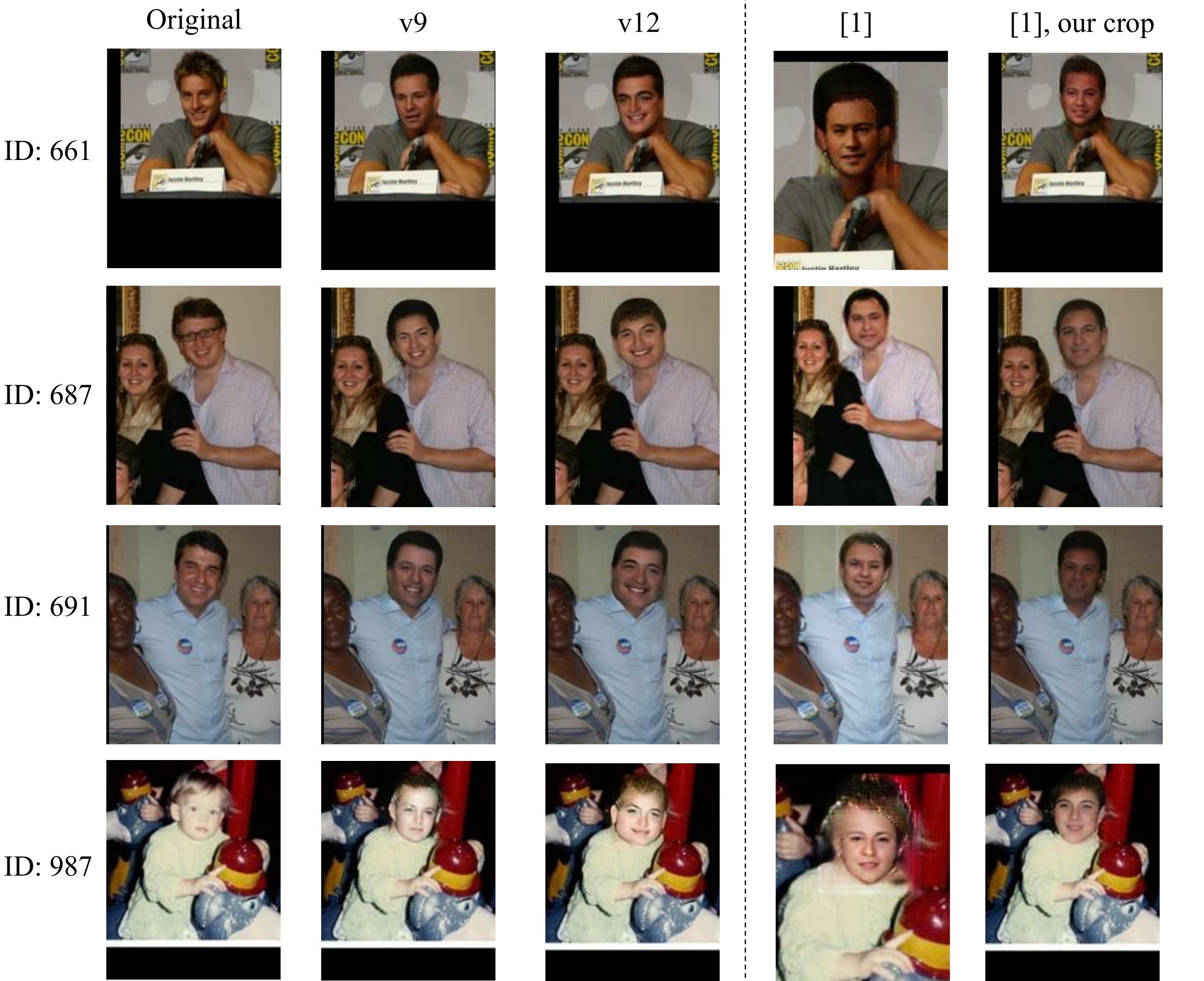}\\
  \vspace{-0.1cm}
  \caption{Result images by methods \textbf{v9} and \textbf{v12}, compared to original images and the results of the \textit{Blackhead} scenario using \textit{PDMDec} landmarks in~\cite{sun2018cvpr}. Note that the image scale difference with~\cite{sun2018cvpr} is because of different cropping methods.}
  \vspace{-0.6cm}
\label{Figure_result_all_comp_compact_blackhair}
\end{figure}

\myparagraph{Comparison to the state-of-the-art.} 
In Table~\ref{Table_quantitative}, we report quantitative evaluation results on different input modalities and in comparison to~\cite{sun2018cvpr}.
We also implement the exact same models of~\cite{sun2018cvpr} on our cropped data for fair comparisons. 
We also compare the visual quality of our results with~\cite{sun2018cvpr}, see Fig.~\ref{Figure_result_all_comp_compact_blackhair}. 

Our best obfuscation rate is achieved by \textbf{v12}. The most comparable method in~\cite{sun2018cvpr} is \emph{Blackhead+PDMDec}, where the input is an image with a fully blacked-out head and the landmarks are generated by \emph{PDMDec}. Comparing \textbf{v12} with it, we achieve $2.6\%$ lower recognition rate ($2.6\%$ higher for confusing machine recognizers) using \texttt{head} features. Our method does even better ($15.3\%$ higher) in fooling human recognizers. In addition, our method has clearly higher image quality in terms of HPS, $0.33$ vs. $0.15$~\cite{sun2018cvpr}. 
Figure~\ref{Figure_result_all_comp_compact_blackhair} shows that our method generates more natural images in terms of consistent skin colors, proper head poses and vivid face expressions. 

\myparagraph{Parametric model versus GAN.}
For the ablation study of our hybrid model, we replace the parametric model in Stage-I with a GAN. We use the same architecture as the stage-II network of \cite{sun2018cvpr}, but without the landmark channel. This is inspired by the regional completion method using GANs \cite{IizukaS017}. 
We consider two comparison scenarios: when the input face is blacked-out (indexed by \textbf{v13}), we compare  with \textbf{v8}; when the head is blacked-out (indexed by \textbf{v14}), we compare with \textbf{v9}. 
We observe that \textbf{v13} results in a lower mask-SSIM score of 0.80 (\textbf{v8} has 0.85) with the same recognition rate of 64.4\%. This means that the GAN generates lower-quality images without performing better obfuscation. \textbf{v14} has a lower recognition rate of 19.7\% vs. \textbf{v9}'s 25.7\%, but its mask-SSIM (image quality) is only 0.23, 0.24 lower than \textbf{v9}'s 0.47. 
If we make use of face replacement (only applicable when using our parametric model-based approach), we are able to achieve a lower recognition rate of 18.1\%, sacrificing only 0.08 in terms of image quality (see HPS of \textbf{v9} and \textbf{v12} in Table \ref{Table_quantitative}).

\begin{figure}[htp]
  \centering
  \includegraphics[width=0.8\linewidth]{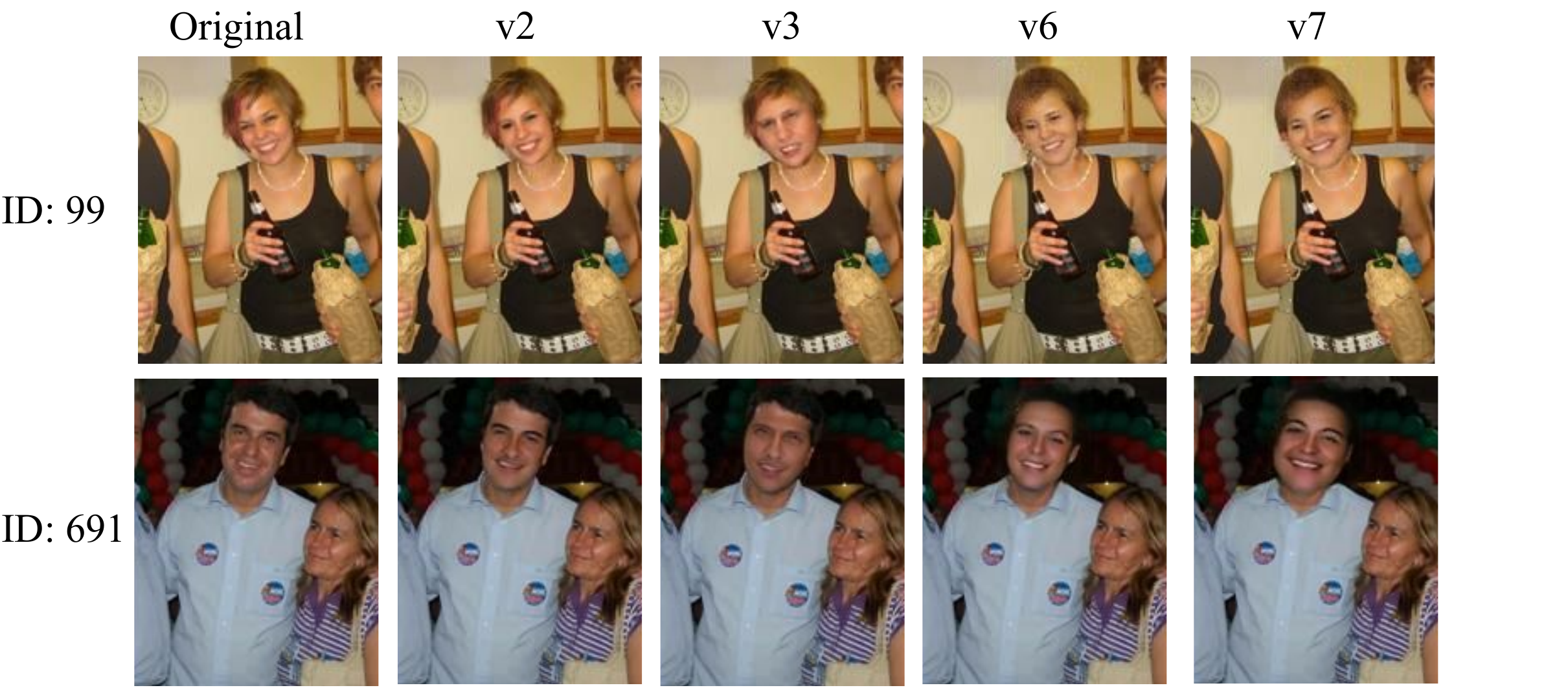}\\
  \vspace{-0.2cm}
  \caption{
  Result images of methods \textbf{v2}, \textbf{v3}, \textbf{v6} and \textbf{v7}, compared to the original images.}
  \vspace{-0.5cm}
\label{Figure_result_all_comp_compact_blur_original}
\end{figure}

\myparagraph{Analysis of different face/hair modalities.}
Table~\ref{Table_quantitative} shows that different modalities of the input yield different levels of obfuscation and image quality.
In general, the image quality is roughly 
correlated to the recognition rate.
With higher level of modification to an image, the identity will be more effectively obfuscated, but the image quality will also deteriorate accordingly.
However, we can observe that the recognition rate drops quicker than the image quality.

It is worth noting that when there is no inpainting on the rendered faces (\textbf{v1}), HPS score is $0.58$, $0.13$ lower than \textbf{v2}, verifying that  rendered faces are less realistic than inpainted ones.
Not surprisingly, the best image quality is achieved by \textbf{v2} which aims to reconstruct the original image without obfuscation.
On top of that, when we use blurred faces in Stage-I (\textbf{v4}),  
the machine recognition rate (\texttt{head}) drops from $70.8\%$ to $59.9\%$.
This indicates that blurring the face region indeed partially conceals the identity information.

When we blur the hair region (\textbf{v6}),
the recognition rate sharply drops to $25.8\%$, which implies that the region around the face contains a large amount of identity information.
When we remove all information from the face and hair regions (\textbf{v9}),
we get an even lower recognition rate of $14.2\%$.

\myparagraph{Face replacement is of great effectiveness.}
We can see from Table~\ref{Table_quantitative} that replacing the face parameters with those of another identity is an effective way of hiding the identity information.
Regardless of the face and hair input modalities, the obfuscation performances on both recognizers are 
significantly improved using \texttt{Replacer15} rendered faces than using \texttt{Own} rendered faces.
Replacing faces from close to far identities also has an obvious impact on the obfuscation effectiveness.
From \textbf{v10} to \textbf{v12} in Table~\ref{Table_quantitative}, 
we can see using \texttt{Replacer8} yields clearly better obfuscation than the \texttt{Replacer1}, e.g., obfuscation for humans gets $25.1\%$ improvement.
This is further evidenced by the comparison between the \texttt{Replacer15} and \texttt{Replacer1}.
Visually, Fig.~\ref{Figure_result_all_comp_compact_blur_original} and Fig.~\ref{Figure_result_all_comp_compact_blackhair} show that replacing the face parameters indeed makes the faces very different.

\begin{figure}[htp]
 \centering
 \includegraphics[width=0.85\linewidth]{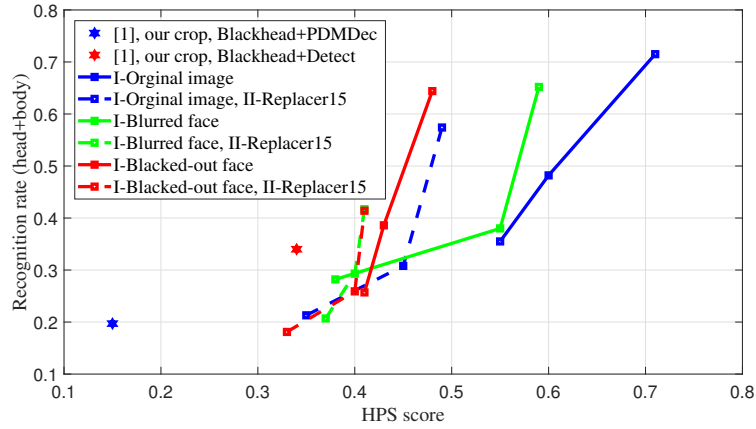}\\
 \vspace{-0.3cm}
 \caption{Scatter curves of different obfuscation methods. HPS scores change along X-axis for different obfuscation levels (\emph{Blacked-out}, \emph{Blurred}, \emph{Original}) 
on hair regions.}
\vspace{-0.4cm}
\label{hps_recog}
\end{figure}  

\myparagraph{Trade-off between image quality and obfuscation.}
Fig.~\ref{hps_recog} shows the machine recognition rate vs.\ image quality plots for different obfuscation methods (some are not in Table~\ref{Table_quantitative} but in supplementary materials).
Points on the curves from left to right are the results of using \emph{Blacked-out}, \emph{Blurred} and \emph{Original} hair inputs for stage-II.

This figure allows users to select the method with the highest image quality given a specified obfuscation threshold.
For example, if a user would like to take the risk of $30\%$ recognizability at most, the highest image quality he/she can get is about $0.45$, corresponding to the middle point on the blue dashed line (the method of \emph{Original image}, \emph{Blurred hair}, \texttt{Replacer15}).
On the other hand, if a user requires the image quality to be at least $0.30$, the best obfuscation possible corresponds to the first point of the red dashed line (the method of \emph{Blacked-out face}, \emph{Blacked-out hair}, \texttt{Replacer15}).
The global coverage of these plots show the selection constrains, such as when a user strictly controls the privacy leaking rate under $20\%$, there are only two applicable methods: \textit{Blackhead}+\textit{PDMDec}~\cite{sun2018cvpr} (image quality is only $0.15$) and ours (\emph{Blacked-out face}, \emph{Blacked-out hair}, \texttt{Replacer15}) where the image quality is higher at $0.33$.

%% file: 5-main_table.tex
\begin{table*}[t]
\centering
\small
\caption{Quantitative results comparing with the state-of-the-art methods~\cite{sun2018cvpr}. 
Image quality: Mask-SSIM and HPS scores (both the higher, the better).
Obfuscation effectiveness: recognition rates of machine recognizers (lower is better) and confusion rates of human recognizers (higher is better). \textbf{v*} simply represents the method in that row.
}

\begin{centering}
\begin{tabular*}{16cm}
{ l l l c c c c c c c c c c}
\multicolumn{3}{c}{Obfuscation method}&\multicolumn{8}{c}{Evaluation}\\
\cmidrule{1-3}\cmidrule{5-11}
&\multicolumn{2}{c}{Stage-II}&& \multicolumn{2}{c}{Image quality} && \multicolumn{2}{c}{Machine}&& Human \\
\cmidrule{2-3}\cmidrule{5-6}\cmidrule{8-9}\cmidrule{11-11}
Stage-I &Hair&Rendered face  && Mask-SSIM & HPS && \texttt{head} & \texttt{body+head}&& confusion \\
\cmidrule{1-3}\cmidrule{5-6}\cmidrule{8-9}\cmidrule{11-11}
Original & -& -&& 1.00 & 0.93 && 85.6\% & 88.3\% &&  \\
\cmidrule{1-3}\cmidrule{5-6}\cmidrule{8-9}
\cite{sun2018cvpr} &\multicolumn{2}{l}{Blackhead+Detect}&& 0.41 & 0.19&& 10.1\%& 21.4\%&& -\\
\cite{sun2018cvpr} &\multicolumn{2}{l}{Blackhead+PDMDec.}&& 0.20 & 0.11&& 5.6\%& 17.4\%&& -\\
\cite{sun2018cvpr}, our crop &\multicolumn{2}{l}{Blackhead+Detect}&& 0.43 & 0.34&& 12.7\%& 24.0\%&& 4.1\% \\
\cite{sun2018cvpr}, our crop &\multicolumn{2}{l}{Blackhead+PDMDec.}&& 0.23 & 0.15&& 9.7\%& 19.7\%&& 20.1\% \\
\cmidrule{1-3}\cmidrule{5-6}\cmidrule{8-9}\cmidrule{11-11}
\textbf{v1}, Original & \multicolumn{2}{l}{Overlay-No-Inpainting}&& 0.75 & 0.58 &&  66.9\% & 68.9\%  &&  \\
\cmidrule{1-3}\cmidrule{5-6}\cmidrule{8-9}
\textbf{v2}, Original&Original&\texttt{Own} && 0.87 & 0.71 && 70.8\% & 71.5\% && - \\
\textbf{v3}, Original&Original&\texttt{Replacer15} && - & 0.49 && 47.6\% & 57.4\% && -   \\
\cmidrule{1-3}\cmidrule{5-6}\cmidrule{8-9}\cmidrule{11-11}
\textbf{v4}, Blurred&Original&\texttt{Own} && 0.86 & 0.59 && 59.9\% & 65.2\% && - \\
\textbf{v5}, Blurred&Original&\texttt{Replacer15} && - & 0.41 && 26.3\% & 41.7\% && -   \\
\textbf{v6}, Blurred&Blurred&\texttt{Own} && 0.55 & 0.55 && 25.8\% & 38.0\% && -   \\
\textbf{v7}, Blurred&Blurred&\texttt{Replacer15} && - & 0.40 && 12.7\% & 29.3\% && -   \\
\cmidrule{1-3}\cmidrule{5-6}\cmidrule{8-9}\cmidrule{11-11}
\textbf{v8}, Blacked&Original&\texttt{Own} && 0.85 &0.60&& 59.3\% & 64.4\% &&-   \\

\textbf{v9}, Blacked&Blacked&\texttt{Own} && 0.47 & 0.41 && 14.2\% & 25.7\% &&  2.9\%  \\
\textbf{v10}, Blacked&Blacked&\texttt{Replacer1} && - & 0.45 && 11.8\% & 23.5\% && 6.2\%  \\
\textbf{v11}, Blacked&Blacked&\texttt{Replacer8} && - & 0.39 && 9.3\% & 22.4\% && 31.3\%  \\
\textbf{v12}, Blacked&Blacked&\texttt{Replacer15} && - & 0.33 && \bf{7.1\%} & \bf{18.1\%} && \bf{35.4\%}   \\
\cmidrule{1-3}\cmidrule{5-6}\cmidrule{8-9}\cmidrule{11-11}
\end{tabular*}
\end{centering}
\vspace{-0.4cm}
\label{Table_quantitative}
\end{table*}

%% file: 9_supple.tex
\clearpage
\setcounter{section}{0}
\renewcommand\thesection{\Alph{section}}
\noindent
{\Large {\textbf{Supplementary materials}}}
\\

\section{Network architectures}
\label{suppsec:arch}

\begin{figure*}[htp]
  \centering
  \includegraphics[width=1\linewidth]{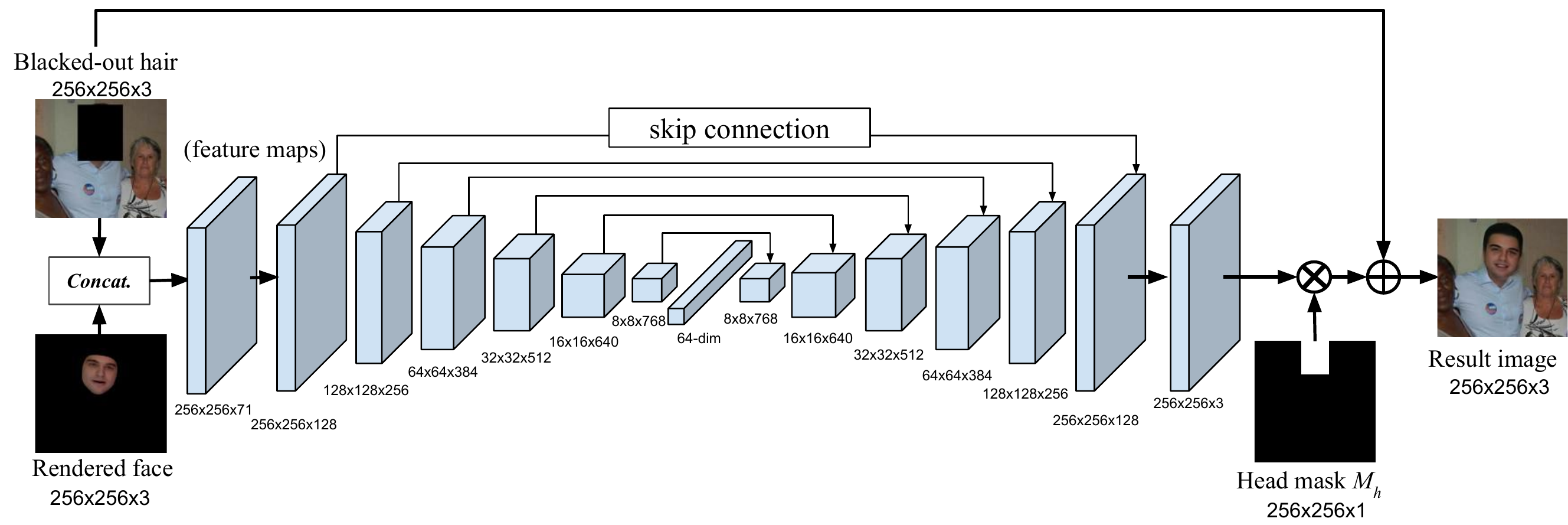}\\
   \vspace{-0.1cm}
     \caption{The network architecture of Head Generator $G$ used in the stage-II.}
  \label{supp_arch_head_G}
\end{figure*}

In Fig.~\ref{supp_arch_head_G}, we present the U-Net architecture for the Head Generator $G$, which corresponds to the Inpainter in Fig.1 of the main paper. 
Note that the output of the deep network is the image (256x256x3) including the body and the head.
In the final layer, the output is cropped based on the head mask and pasted onto the obfuscated image (one of the inputs). 
Therefore, only the head region can provide any feedback during back-propagation. This follows from~\cite{sun2018cvpr}.

\section{Implementation details}
\label{suppsec:implementation}
For the Stage-I network (Section 3.1 in the main paper), AlexNet is used as the encoder (``Conv Encoder'' in the Fig.1 of the main paper). We use AdaDelta~\cite{zeiler2012adadelta} ($200$k iterations) to optimize the weights of the network with a batch size of 5 and  a learning rate of $10^{-3}$.

In the Inpainter (Section 3.2 in the main paper), the Head Generator is trained using the Adam optimizer~\cite{Adam} with $\lambda_H = 1000$ (in the main paper Eq. (7)). Initial learning rates (for both generator $G$ and discriminator $D$) are $2\times 10^{-5}$, which decays to half every $5,000$ iterations.
The batch size is set to $6$; optimization stops after $10,000$ iterations; each iteration consists of $5$ and $1$ parameter updates for the generator and the discriminator respectively. It takes around $9$ epochs for training the generator and around $2$ epochs for training the discriminator.

To prepare a $256 \times 256$ body crop (Section 5.1 in the main paper), keeping the ratio of the head (width/height) unchanged, we first resize the original image such that the head height is $1/4$ of $256$. Then, we crop a $3$ head weight $\times$ $4$ head height region from the input image,
making the head lying in the upper middle region of the crop. We zero-pad the image if its dimensions are smaller than the crop size, thereby obtaining the final crop with the desired $256 \times 256$ size.

\input{supple_main_table}
\input{9_supple_results}


%% file: supple_main_table.tex
\begin{table*}[t]
\centering
\small
\caption{Quantitative results comparing with the state-of-the-art methods~\cite{sun2018cvpr}. 
Image quality: Mask-SSIM, SSIM and HPS scores (both the higher, the better).
Obfuscation effectiveness: recognition rates of machine recognizers (lower is better). \textbf{v*} simply represents the method in that row, noting that supplementary methods are numbered after \textbf{v12} according to the Table 1 of our main paper. To save space, we use some abbreviations of input data as Rendered.=Rendered Face, Orig.=Original, Blu.=Blurred, Bla.=Blacked and Overlay-No-Inp.=Overlay-No-Inpainting, while full names were used in the Table 1 of our main paper.
}
\vspace{-0.1cm}
\begin{centering}
\begin{tabular*}{16cm}
{ l l l c c c c c c c c c c c}
\multicolumn{3}{c}{Obfuscation method}&\multicolumn{8}{c}{Evaluation}\\
\cmidrule{1-3}\cmidrule{5-12}
&\multicolumn{2}{c}{Stage-II}&& \multicolumn{2}{c}{Image quality}&& \multicolumn{2}{c}{Google Net}&& \multicolumn{2}{c}{Alex Net}  \\
\cmidrule{2-3}\cmidrule{5-6}\cmidrule{8-9} \cmidrule{11-12}
Stage-I &Hair&Rendered.  && Mask-SSIM& SSIM && \texttt{head} & \texttt{body+head}&&  \texttt{head} & \texttt{body+head}\\
\cmidrule{1-3}\cmidrule{5-6}\cmidrule{8-9} \cmidrule{11-12}
Orig. & -& -&& 1.00 &1.00&& 85.6\% & 88.3\% && 81.6\% & 85.3\% \\
\cmidrule{1-3}\cmidrule{5-6}\cmidrule{8-9} \cmidrule{11-12}
\cite{sun2018cvpr} &\multicolumn{2}{l}{Blu.+Detect}&& 0.68 &0.96&& 43.7\%& 51.7\%&&49.0\% & 48.9\% \\
\cite{sun2018cvpr} &\multicolumn{2}{l}{Blu.+PDMDec.}&& 0.59 &0.95&& 37.9\%& 49.1\%&& 45.1\% & 45.6\% \\
\cite{sun2018cvpr} &\multicolumn{2}{l}{Bla.+Detect}&& 0.41 &0.90&& 10.1\%& 21.4\%&&11.4\% & 20.5\% \\
\cite{sun2018cvpr} &\multicolumn{2}{l}{Bla.+PDMDec.}&& 0.20 &0.86&& 5.6\%& 17.4\%&&7.4\% & 16.6\% \\
\cite{sun2018cvpr}, our crop &\multicolumn{2}{l}{Blu.+Detect}&& 0.64 &0.98&& 40.5\%& 47.8\%&&43.6\% & 43.2\%  \\
\cite{sun2018cvpr}, our crop &\multicolumn{2}{l}{Blu.+PDMDec.}&& 0.47 &0.97&& 30.6\%& 38.6\%&&35.4\% & 37.0\%  \\
\cite{sun2018cvpr}, our crop &\multicolumn{2}{l}{Bla.+Detect}&& 0.43 &0.97&& 12.7\%& 24.0\%&&15.1\% & 23.4\% \\
\cite{sun2018cvpr}, our crop &\multicolumn{2}{l}{Bla.+PDMDec.}&& 0.23 &0.96&& 9.7\%& 19.7\%&&10.5\% & 19.2\% \\
\cmidrule{1-3}\cmidrule{5-6}\cmidrule{8-9} \cmidrule{11-12}
\textbf{v1}, Orig. & \multicolumn{2}{l}{Overlay-No-Inp.}&& 0.75  &0.96&&  66.9\% & 68.9\%  && 64.0\% & 54.9\%   \\
\cmidrule{1-3}\cmidrule{5-6}\cmidrule{8-9} \cmidrule{11-12}
\textbf{v2}, Orig.&Orig.&\texttt{Own} && 0.87  &0.99&& 70.8\% & 71.5\% &&66.6\% & 58.3\%  \\
\textbf{v3}, Orig.&Orig.&\texttt{Replacer15} && - &-&& 47.6\% & 57.4\% &&45.1\% & 47.9\%   \\
\textbf{v13}, Orig.&Blu.&\texttt{Own} && 0.58  &0.98&& 36.6\% & 48.2\% &&42.4\% & 43.2\%  \\
\textbf{v14}, Orig.&Blu.&\texttt{Replacer15} && - &-&& 18.0\% & 30.8\% &&22.9\% & 30.2\%   \\
\textbf{v15}, Orig.&Bla.&\texttt{Own} && 0.50 &0.97&& 22.5\% & 35.5\% &&30.3\% & 33.9\%  \\
\textbf{v16}, Orig.&Bla.&\texttt{Replacer15} && -  &-&& 7.1\% & 21.3\% &&13.2\% & 21.5\%   \\
\cmidrule{1-3}\cmidrule{5-6}\cmidrule{8-9} \cmidrule{11-12}
\textbf{v4}, Blu.&Orig.&\texttt{Own} && 0.86  &0.99&& 59.9\% & 65.2\% &&57.8\% & 52.0\%  \\
\textbf{v5}, Blu.&Orig.&\texttt{Replacer15} && - &-&& 26.3\% & 41.7\% &&24.1\% & 31.8\%    \\
\textbf{v6}, Blu.&Blu.&\texttt{Own} && 0.55  &0.98&& 25.8\% & 38.0\% &&28.2\% & 33.5\%    \\
\textbf{v7}, Blu.&Blu.&\texttt{Replacer15} && -  &-&& 12.7\% & 29.3\% &&14.8\% & 23.3\%    \\
\textbf{v17}, Blu.&Bla.&\texttt{Own} && 0.44  &0.97&& 15.7\% & 28.2\% &&19.7\% & 25.7\%   \\
\textbf{v18}, Blu.&Bla.&\texttt{Replacer15} && -  &-&& 7.2\% & 20.7\% &&9.9\% & 18.9\%   \\
\cmidrule{1-3}\cmidrule{5-6}\cmidrule{8-9} \cmidrule{11-12}
\textbf{v8}, Bla.&Orig.&\texttt{Own} && 0.85 &0.99&& 59.3\% & 64.4\% &&54.8\% & 49.1\%   \\
\textbf{v19}, Bla.&Orig.&\texttt{Replacer15} && - &-&& 27.0\% & 41.4\% &&25.0\% & 31.3\% \\
\textbf{v20}, Bla.&Blu.&\texttt{Own} && 0.53 &0.98&& 28.1\% & 38.6\% &&31.0\% & 34.7\% \\
\textbf{v21}, Bla.&Blu.&\texttt{Replacer15} && - &-&& 11.2\% & 25.9\% &&14.7\% & 22.1\%  \\
\textbf{v9}, Bla.&Bla.&\texttt{Own} && 0.47 &0.97&& 14.2\% & 25.7\% &&19.1\% & 24.4\%  \\
\textbf{v12}, Bla.&Bla.&\texttt{Replacer15} && - &-&& \bf{7.1\%} & \bf{18.1\%} &&\bf{9.7\%} & \bf{16.5\%}  \\
\cmidrule{1-3}\cmidrule{5-6}\cmidrule{8-9} \cmidrule{11-12}
\end{tabular*}
\end{centering}
\vspace{-0.4cm}
\label{Table_quantitative_supp}
\end{table*}

%% file: 9_supple_results.tex
\section{Obfuscation performance against AlexNet}
\label{suppsec:alexnet}
In the experiments provided in the main paper, we focused on the obfuscation performance using a GoogleNet-based recognizer. However, as we have mentioned, our approach is \emph{target-generic}: it is expected to work against a generic system.

Therefore, in this section, we additionally present the obfuscation performance with respect to an AlexNet-based recognizer.
Following the same ``feature extraction - SVM prediction'' framework as in the main paper, we replace the feature extractor with AlexNet. 
Table~\ref{Table_quantitative_supp} shows the quantitative comparisons between GoogleNet and AlexNet recognizers on different versions of our approach. 
Note that in this table, \textbf{v1}$\sim$\textbf{12} are the same representative versions as shown in Table 1 of the main paper. All other versions are also added here. Note that \textbf{v13} and \textbf{v14} are indexing two hybrid models, different with the GAN models for stage-I ablation study in the main paper.

Some recognition rate differences exit between the two recognizers. First of all, on original (ground truth) images, AlexNet performs worse than GoogleNet ($81.6\%<85.6\%$). 
On images generated by our method (\textbf{v1}$\sim$\textbf{21}), AlexNet performs similarly when using head features, achieving a higher recognition rate for 12 input modalities out of 21, compared to GoogleNet. However, while using head+body features, GoogleNet recognition rates are higher for 18 different input modalities.
The possible reason could be that the 1024-dimensional GoogleNet features are more compact than the AlexNet features, which are 4096-dimensional. From the discriminative head images, less compact features can extract more information in the additional feature dimensions. On the other hand, concatenation of features from the noisy body images could reduce the final recognition rates.

\section{Visualization results}
\label{suppsec:vis}

In this section, we show visualization results using different modalities (\textbf{v2} to \textbf{v21}), corresponding to Table~\ref{Table_quantitative_supp}. 

\begin{figure}[htp]
  \centering
  \includegraphics[width=0.99\linewidth]{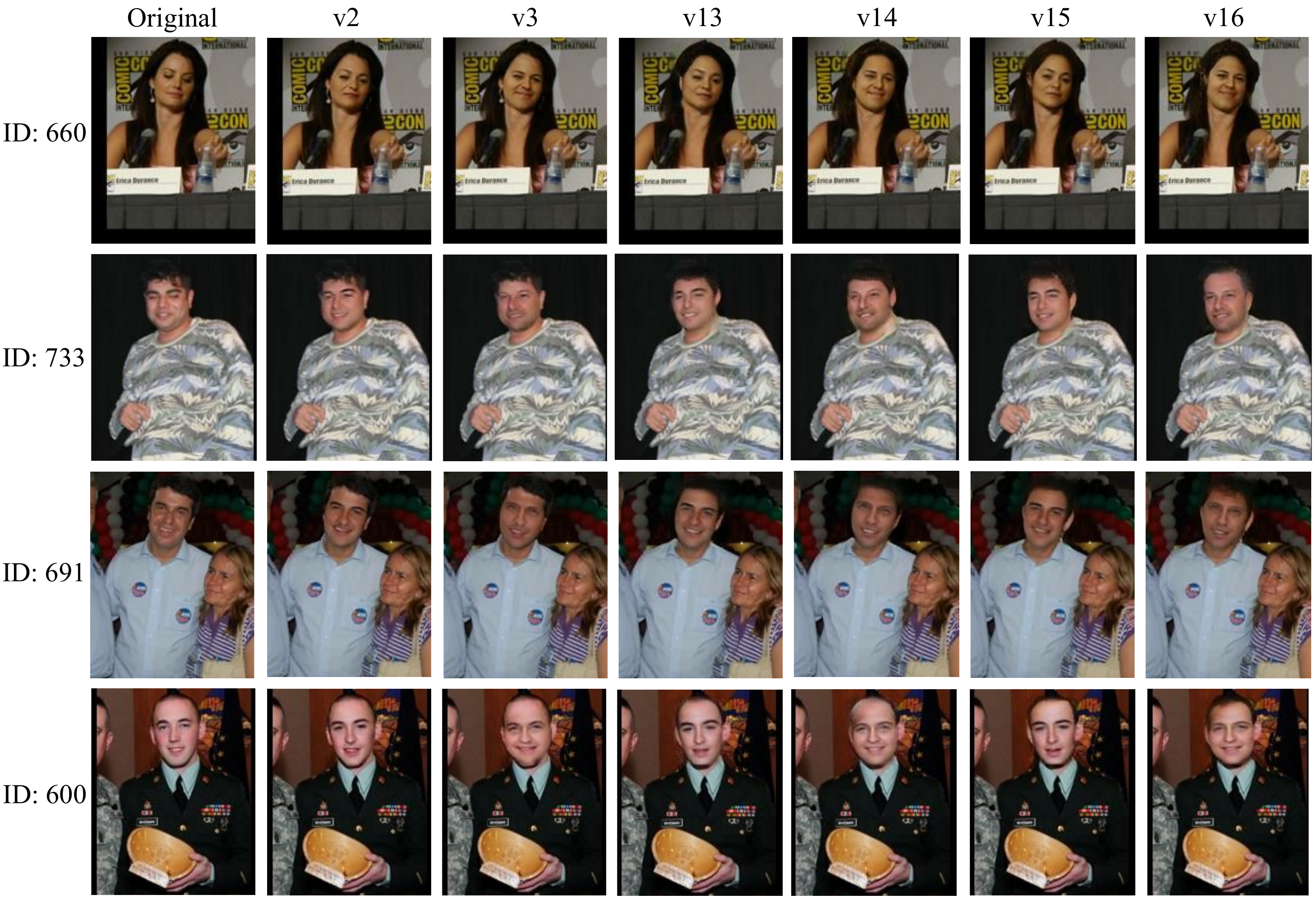}\\
  \caption{
  Result images of methods \textbf{v2} to \textbf{v15} in the block named ``Original in the Stage-I'' in Table~\ref{Table_quantitative_supp}, compared to original images.}
\label{Figure_result_all_comp_compact_blur_smaller_cropped_supple}
\end{figure}

\begin{figure}[htp]
  \centering
  \includegraphics[width=0.99\linewidth]{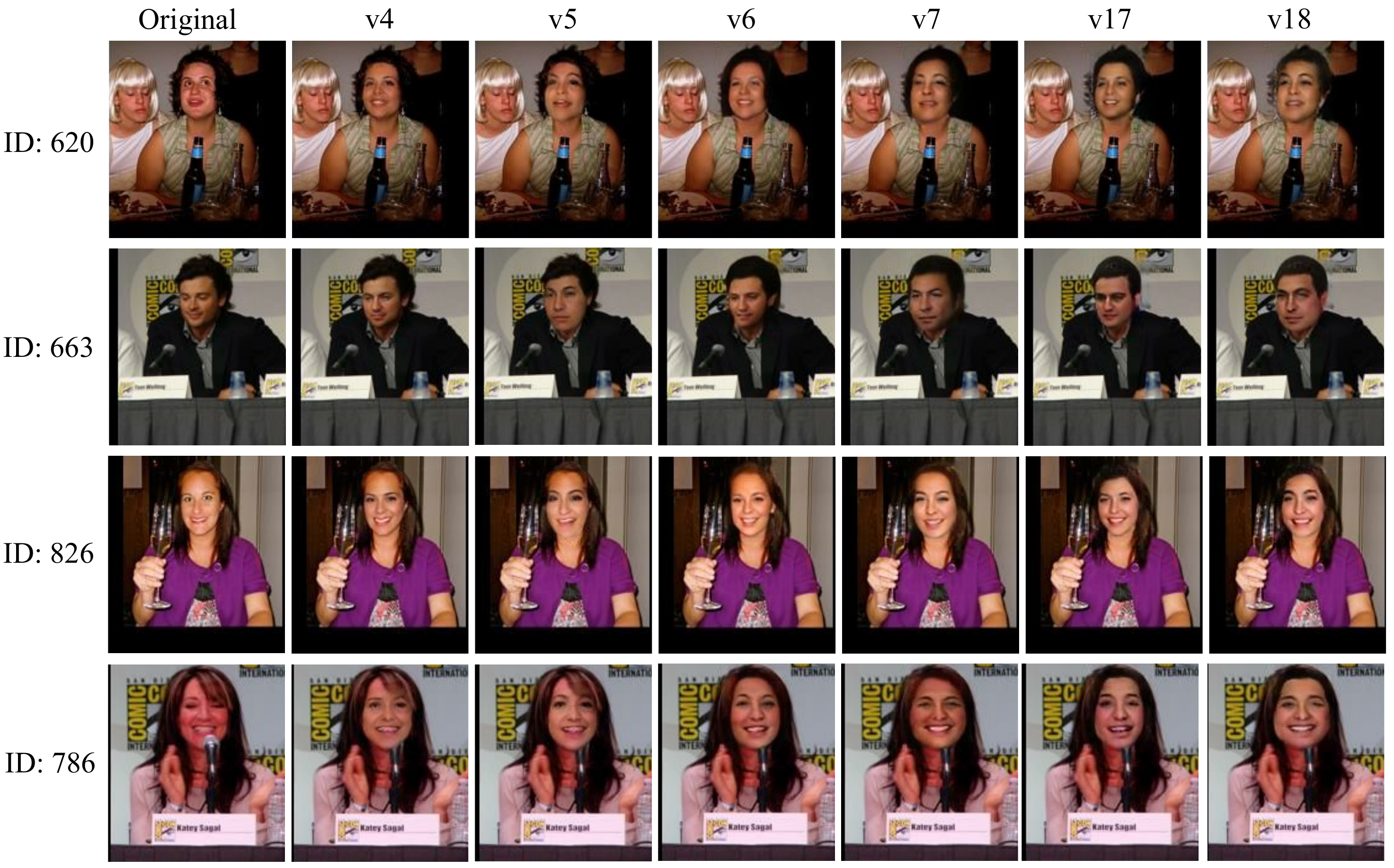}\\
  \caption{Result images of methods \textbf{v4} to \textbf{v18} in the block named ``Blurred in the Stage-I'' in Table~\ref{Table_quantitative_supp}, compared to original images.}
\label{Figure_result_all_comp_compact_blurblock_cropped_supple}
\end{figure}

\begin{figure}[htp]
  \centering
  \includegraphics[width=0.96\linewidth]{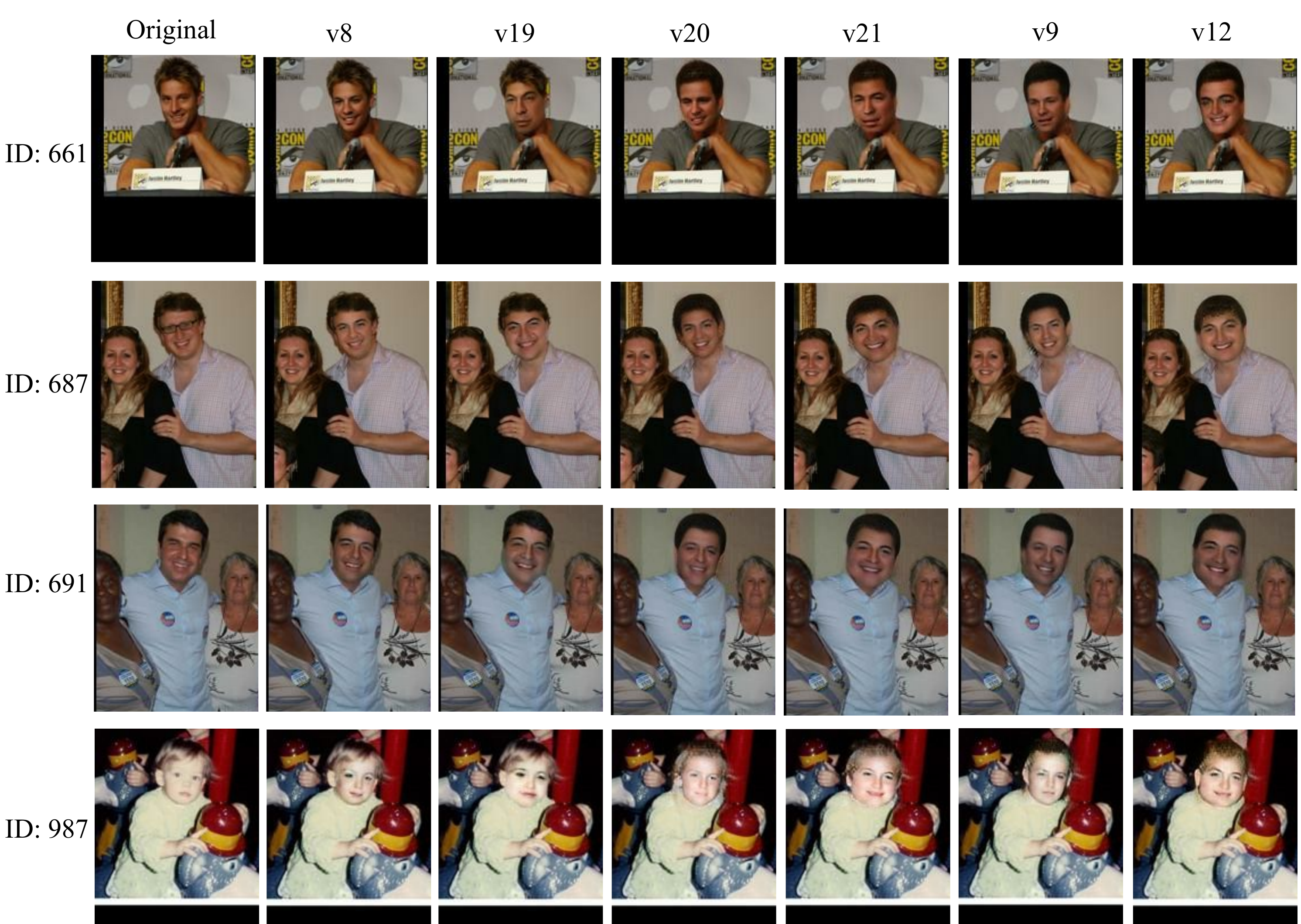}\\
  \caption{Result images of methods \textbf{v8} to \textbf{v12} in the block named ``Blacked in the Stage-I'' in Table~\ref{Table_quantitative_supp}, compared to original images.}
\label{Figure_result_all_comp_compact_blackhair_smaller_cropped_supple}
\end{figure}

Respectively in Figure~\ref{Figure_result_all_comp_compact_blur_smaller_cropped_supple}, Figure~\ref{Figure_result_all_comp_compact_blurblock_cropped_supple} and Figure~\ref{Figure_result_all_comp_compact_blackhair_smaller_cropped_supple}, we show results with rendered faces from Original images, Blurred face images and Blacked-out face images. Note that, the results are consistently cropped to have small zero-padded regions.
In most cases, the best visual quality is achieved in the second column which uses Original hair images. The largest visual differences compared to the original faces are visible in the last column when the rendered faces are replaced and the hair regions are entirely obfuscated by blacking-out.

